\documentclass[letterpaper,10pt,conference]{ieeeconf}

\ifdefined\pdfminorversion
\fi

\IEEEoverridecommandlockouts
\usepackage{amsmath,amssymb,bm}
\usepackage{booktabs,array,tabularx}
\usepackage{cite}
\usepackage{graphicx}
\usepackage{multirow}
\usepackage{url}
\usepackage{etoolbox}
\usepackage{eso-pic}
\usepackage[table]{xcolor}
\usepackage[hidelinks,unicode,bookmarksnumbered]{hyperref}

\newif\ifanonymous
\anonymousfalse
\newcommand{\papertitle}{EgoDepth-Bench: Benchmarking\\
Metric Depth for Egocentric Manipulation}

\newcommand{\publicaffiliation}{GenRobot}
\newcommand{\finalauthors}{%
Yuan Gao\textsuperscript{*}, Dongchen Wei\textsuperscript{\textdagger}, Xinyang Lin, Shuhua Cao, Xueliang Guo, Zhuohan Li, Yihao Wang
\thanks{\textsuperscript{*}Corresponding author: \texttt{gaoyuan100@outlook.com}.\quad \textsuperscript{\textdagger}Project lead.}%
}

\title{\LARGE \bf \papertitle}

\definecolor{projectlink}{HTML}{275C8A}
\newcommand*{\projecturl}{https://egodepth-bench.github.io}
\author{\finalauthors\\[0.15em]
  \ifdefempty{\publicaffiliation}{}{{\small\publicaffiliation}\\[0.1em]}%
  {\normalsize\fontfamily{lmtt}\selectfont
   \href{\projecturl/}{\textcolor{projectlink}{\projecturl}}}}
\ifdefempty{\publicaffiliation}{}{\IEEEaftertitletext{\vspace{-\baselineskip}}}
\hypersetup{
  pdftitle={EgoDepth-Bench: Benchmarking Metric Depth for Egocentric Manipulation},
  pdfauthor={Yuan Gao; Dongchen Wei; Xinyang Lin; Shuhua Cao; Xueliang Guo; Zhuohan Li; Yihao Wang},
  pdfsubject={Preprint}
}

\begin{document}
\widowpenalty=10000
\AddToShipoutPictureFG*{%
  \AtPageLowerLeft{\raisebox{20pt}[0pt][0pt]{%
    \makebox[\paperwidth]{\parbox[b]{\textwidth}{\centering\footnotesize
      This work has been submitted to the IEEE for possible publication.\\
      Copyright may be transferred without notice, after which this version may no longer be accessible.}}}}}
\maketitle
\thispagestyle{empty}
\pagestyle{empty}

\begin{abstract}
We introduce EgoDepth-Bench, comprising 941,233 synchronized samples collected across 37 distinct physical environments. Each sample combines six calibrated RGB views with metric depth in the native fisheye view, reconstructed from five active-infrared stereo modules.
A calibrated ball-bar confirms millimeter-level geometric accuracy, with a median error of 1.27 mm on a known sphere spacing. Environment-held-out splits enable evaluation of generalization to unseen environments. Across monocular, stereo, and multi-view models, the benchmark measures both depth accuracy and prediction coverage, including on hands and objects. Fine-tuning MapAnything on EgoDepth-Bench reduces metric AbsRel from 0.708 to 0.091 and scale-aligned AbsRel from 0.327 to 0.072 on unseen environments.
These results demonstrate that EgoDepth-Bench provides effective supervision and enables rigorous evaluation of metric and relative depth estimation in egocentric manipulation. The project page provides the evaluation code and interactive RGB-depth sequences.

\end{abstract}

\section{Introduction}
\label{sec:introduction}

Egocentric demonstrations provide a rich source of training data for robot
manipulation~\cite{kareer2025egomimic}. Recovering the metric geometry of
hands, manipulated objects, and the surrounding workspace can further support learning from these demonstrations~\cite{shridhar2022peract,ze2024dp3}.
However, depth estimation in egocentric manipulation is particularly challenging: head motion is coupled with independent hand and object motion, occlusions change continuously, and fisheye cameras with a wide field of view introduce strong viewing-angle variation. Evaluating depth in this setting therefore requires metric ground truth on dynamic surfaces, together with measurements that explicitly characterize the interacting hands and objects.

Existing datasets address only subsets of these requirements.
HOI4D captures real egocentric manipulation using perspective RGB-D cameras~\cite{liu2022hoi4d}, while WideDepth evaluates fisheye depth using imagery rendered from static laser-scanned scenes~\cite{indyk2026widedepth}.
Aria Digital Twin (ADT) provides real egocentric interaction and native-fisheye depth rendered from registered scene and object models, but is limited to two physical environments~\cite{pan2023adt}.
Thus, existing resources do not jointly provide native-fisheye metric depth for real dynamic manipulation, broad coverage across physical environments, held-out-location evaluation, and interaction-specific depth analysis.

\begin{figure}[t]
  \centering
  \includegraphics[width=\columnwidth]{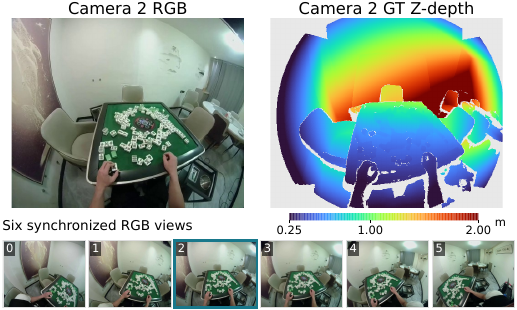}
  \caption{EgoDepth-Bench records head-worn interaction at 37 physical sites.
  Six synchronized RGB views accompany ground-truth Z-depth aligned with camera~2's
  fisheye image. Gray pixels have no valid ground truth; the displayed color scale saturates
  outside 0.25--2\,m. This training-set example illustrates the dataset.}
  \label{fig:teaser}
\end{figure}

We introduce EgoDepth-Bench, a metric depth dataset and benchmark of
real egocentric manipulation. A head-mounted camera rig records 941,233 synchronized samples across 37 distinct physical environments (Fig.~\ref{fig:teaser}).
Each sample contains six synchronized and calibrated RGB views, enabling monocular, stereo, and multi-view evaluation on the same interactions. Disjoint training, validation, and test environments support evaluation of generalization to unseen physical locations.

Metric Z-depth for the target fisheye camera is reconstructed from five active-infrared stereo modules using learned stereo estimation followed by calibrated multi-view fusion in the target camera's native fisheye image coordinates. Unlike approaches based on prebuilt room or object models, this reconstruction captures dynamic hands and manipulated objects directly. We independently validate its metric accuracy using a calibrated ball-bar across depth ranges, viewing directions, and motion conditions, obtaining millimeter-level geometric agreement (Section~\ref{sec:validation}).

We evaluate 18 publicly available depth models under explicit input and scale conventions. The benchmark measures both depth accuracy and prediction coverage over the full image and separately on hand and object regions. Joint depth-angle analysis and matched comparisons across region and motion conditions characterize how performance varies with interaction geometry. Controlled experiments further isolate the effects of input views, camera-rig geometry, and training on EgoDepth-Bench.

Our evaluation shows that low depth error alone can obscure incomplete predictions, particularly on manipulated objects, motivating the joint use of accuracy and coverage. Training on EgoDepth-Bench substantially improves both metric and relative depth estimation at held-out sites. With rig geometry fixed, fine-tuning MapAnything~\cite{keetha2025mapanything} reduces test AbsRel from 0.708 to 0.091, with substantial gains remaining after per-frame scale alignment.

Our contributions are:
\begin{enumerate}
  \item A real egocentric manipulation dataset spanning 37 distinct physical environments, with six synchronized RGB views, native-fisheye metric Z-depth, and independent ball-bar validation of reconstructed geometry.
  \item A depth benchmark that jointly evaluates accuracy and prediction coverage, including dedicated measurements on hands and manipulated objects and diagnostics over viewing geometry and motion.
  \item A systematic evaluation of public monocular, stereo, and multi-view models, together with controlled studies of input views, rig geometry, and cross-environment generalization and adaptation.
\end{enumerate}

\section{Related Work}
\label{sec:related_work}

\subsection{Depth Benchmarks}

NYUv2, KITTI Depth, and DIODE evaluate depth in indoor RGB-D scenes,
driving, and indoor--outdoor imagery, respectively
~\cite{silberman2012indoor,uhrig2017sparsity,vasiljevic2019diode}.
Bonn RGB-D Dynamic includes real moving scenes and object manipulation,
while Dynamic Replica provides synthetic stereo sequences with moving
people and animals~\cite{palazzolo2019refusion,karaev2023dynamicstereo}.

Camera geometry and capture viewpoint motivate more targeted benchmarks.
ScanNet++ registers fisheye photographs to laser-scanned
geometry~\cite{yeshwanthliu2023scannetpp}. WideDepth renders its indoor
fisheye benchmark from scans and provides separate real outdoor training
data with LiDAR depth~\cite{indyk2026widedepth}. AerialMetric combines
targeted evaluation of UAV imaging conditions with model
adaptation~\cite{song2026aerialmetric}. EgoDepth-Bench examines real
head-worn manipulation through local depth diagnostics and adaptation
across physical sites.

\subsection{Egocentric Interaction Geometry}

HOI4D uses head-mounted
RGB-D sensors~\cite{liu2022hoi4d}. ADT records one RGB and two monochrome
views and renders aligned fisheye depth from registered
scene and object models~\cite{pan2023adt}.
HOT3D and SHOW3D provide hand and object poses and meshes
for interaction geometry~\cite{banerjee2025hot3d,rim2026show3d}.

The Ego-1K research release offers 12 rectified RGB views and evaluates
stereo consistency~\cite{lee2026ego1k}. EgoLive reconstructs depth from
rectified RGB stereo pairs and validates it against physical
targets~\cite{li2026egolivelargescaleegocentricdataset}; Xperience-10M
provides stereo-derived depth for egocentric
interaction~\cite{ropedia2026xperience}.

These resources already support depth evaluation: EgoM2P evaluates video
depth on HOI4D with sequence-level scale and shift
alignment~\cite{li2025egom2p}, and UniK3D reports zero-shot results on
ADT~\cite{piccinelli2025unik3d}. EgoDepth-Bench reconstructs native fisheye ground-truth depth from
dedicated active-IR pairs, retaining six RGB views as model inputs and
separating training and evaluation across physical sites
(Table~\ref{tab:benchmark_comparison}).

\subsection{Depth Models and Geometric Inputs}

Depth Anything~v2 and Marigold predict relative depth
~\cite{yang2024depthanythingv2,ke2024marigold}; Metric3D~v2 and
UniDepth~v2 estimate metric depth
~\cite{hu2024metric3dv2,piccinelli2025unidepthv2}.
Depth Any Camera and UniK3D support a broader range of camera
projections~\cite{guo2025dac,piccinelli2025unik3d}.
FoundationStereo uses rectified image pairs~\cite{wen2025foundationstereo},
while VGGT and MapAnything recover geometry from multiple images;
MapAnything also accepts supplied calibration
~\cite{wang2025vggt,keetha2025mapanything}.
Our evaluation groups models by these input and scale assumptions,
separating metric accuracy from accuracy after alignment to ground truth.

\begin{table*}[t]
\definecolor{benchaccent}{RGB}{35,71,91}
\centering
\fontsize{8}{9.5}\selectfont
\setlength{\tabcolsep}{2pt}
\renewcommand{\arraystretch}{1.18}
\renewcommand{\tabularxcolumn}[1]{m{#1}}
\caption{Comparison of depth datasets and egocentric interaction resources.}
\label{tab:benchmark_comparison}
\begin{tabularx}{\textwidth}{@{}>{\raggedright\arraybackslash}m{100pt}>{\centering\arraybackslash}m{30pt}>{\centering\arraybackslash}m{28pt}>{\centering\arraybackslash}m{25pt}>{\centering\arraybackslash}m{27pt}>{\centering\arraybackslash}m{38pt}>{\centering\arraybackslash}m{33pt}>{\raggedright\arraybackslash}X>{\raggedright\arraybackslash}m{64pt}>{\raggedright\arraybackslash}m{42pt}@{}}
\toprule
& \multicolumn{3}{c}{Capture conditions} & \multicolumn{2}{c}{Visual observations} & \multicolumn{2}{c}{Depth annotations} & \multicolumn{2}{c}{Dataset scale} \\
\cmidrule(lr){2-4}\cmidrule(lr){5-6}\cmidrule(lr){7-8}\cmidrule(l){9-10}
Dataset & Type & \shortstack{Head-\\worn} & Manip. & \shortstack{RGB\\views} & \shortstack{RGB\\HFoV} & \shortstack{Fisheye\\depth} & Construction & Reported size & \shortstack[l]{Scenes /\\sites} \\
\midrule
\multicolumn{10}{@{}l}{\textit{General and dynamic depth datasets}} \\
NYUv2 (labeled)~\cite{silberman2012indoor} & Real & -- & -- & 1 & NR & -- & RGB-D & 1.4K pairs & 464 scenes \\
KITTI Depth~\cite{uhrig2017sparsity} & Real & -- & -- & 2 & NR & -- & LiDAR & $>$93K maps & NR \\
DIODE (train/val)~\cite{vasiljevic2019diode} & Real & -- & -- & 1 & $60^\circ$ & -- & LiDAR & 26.2K pairs & 25 scenes \\
\shortstack[l]{Bonn RGB-D\\Dynamic~\cite{palazzolo2019refusion}} & Real & -- & $\checkmark$ & 1 & NR & -- & RGB-D & 26 seq. & NR \\
Dynamic Replica~\cite{karaev2023dynamicstereo} & Synth. & -- & -- & 2 & NR & -- & Scene rendering & \shortstack[l]{169.2K\\stereo pairs} & NR \\
\midrule
\multicolumn{10}{@{}l}{\textit{Wide-angle depth datasets}} \\
\shortstack[l]{ScanNet++\\(DSLR, v1)~\cite{yeshwanthliu2023scannetpp}} & Real & -- & -- & 1 & NR & $\checkmark$ & Scan rendering & 280K images & 460 scenes \\
WideDepth (bench.)~\cite{indyk2026widedepth} & \shortstack{Scan\\render} & -- & -- & 2 & $120$--$195^\circ$ & $\checkmark$ & Scan rendering & \shortstack[l]{$\sim$5K\\stereo pairs} & 101 scenes \\
WideDepth (train)~\cite{indyk2026widedepth} & Real & -- & -- & 2 & $\sim180^\circ$ & $\checkmark$ & LiDAR & 18K samples & NR \\
\midrule
\multicolumn{10}{@{}l}{\textit{Egocentric interaction datasets}} \\
HOI4D~\cite{liu2022hoi4d} & Real & $\checkmark$ & $\checkmark$ & 2$^a$ & NR & -- & RGB-D & 2.4M frames & 610 rooms \\
Aria Digital Twin~\cite{pan2023adt} & Real & $\checkmark$ & $\checkmark$ & 1 & $110^\circ$ & $\checkmark$ & \shortstack[l]{Tracked-asset\\rendering} & 200 seq. & 2 sites \\
Ego-1K (research)~\cite{lee2026ego1k} & Real & $\checkmark$ & $\checkmark$ & 12 & $120^\circ$ & -- & NR & 491K frames & NR \\
EgoLive~\cite{li2026egolivelargescaleegocentricdataset} & Real & $\checkmark$ & $\checkmark$ & 2 & $130^\circ$ & NR & RGB stereo & 1680 h & NR \\
\midrule
\rowcolor{benchaccent!9}
\textbf{EgoDepth-Bench} & Real & $\checkmark$ & $\checkmark$ & 6 & $\sim163^\circ$ & $\checkmark$ & \shortstack[l]{5$\times$ active-IR\\stereo + fusion} & 941K pairs & 37 sites \\
\bottomrule
\end{tabularx}
\par\smallskip
\parbox{\textwidth}{\fontsize{7.5}{9}\selectfont
Manip.: hand--object interaction with object motion. Fisheye depth: depth annotations in the native fisheye projection; coverage may be partial.
HFoV: horizontal RGB field of view. NR: not specified in the cited source; --: absent.
Scale entries retain their stated units.
$^a$HOI4D captures two synchronized RGB views; released clips are organized per sensor.}
\end{table*}

\section{The EgoDepth-Bench Dataset}
\label{sec:benchmark}
\label{sec:dataset}

EgoDepth-Bench pairs six synchronized RGB views of head-worn interaction
with metric ground-truth Z-depth pixel-aligned with camera~2's fisheye RGB image, supporting
monocular, stereo, and multi-view comparisons on the same observations.

\subsection{Data Collection and Coverage}
\label{sec:data_collection}

Five participants performed freely chosen activities such as folding
clothes, tidying tables, cleaning, and mopping, without prescribed head
or hand movements. Their 1,155 recordings span 37 physical sites and
8.72 hours. Table~\ref{tab:dataset_statistics} summarizes the environments
and data partitions.

The head-mounted rig combines six global-shutter RGB cameras with five
Gemini~336L active-infrared (IR) stereo modules whose views overlap in the
workspace (Fig.~\ref{fig:gt_pipeline}). RGB is recorded at $1600\times1296$.
Camera~2 has a calibrated RGB field of view of approximately $163^\circ$
horizontally and $136^\circ$ vertically, exceeding $200^\circ$ diagonally.
Each IR module records
$848\times480$ stereo pairs with a nominal 95\,mm baseline.
A shared hardware trigger synchronizes RGB and IR acquisition at 30\,Hz.

The dataset contains 941,233 synchronized timestamps, each with six RGB
images and one camera-2 ground-truth depth map, totaling 5,647,398 RGB
images. RGB images and depth maps are provided at $512\times416$ and approximately 30\,Hz, accompanied by
calibration at this resolution and original timestamps.

\begin{table}[t]
\caption{Dataset coverage by environment and site-disjoint splits.}
\label{tab:dataset_statistics}
\centering
\small
\setlength{\tabcolsep}{4pt}
\begin{tabular*}{\columnwidth}{@{\extracolsep{\fill}}lrrrr@{}}
\toprule
\multirow{2}{*}{Environment} & \multicolumn{3}{c}{Sites}
  & \multirow{2}{*}{Recordings} \\
\cmidrule(lr){2-4}
& Train & Val. & Test & \\
\midrule
Domestic & 5 & 0 & 2 & 220 \\
Retail & 7 & 1 & 1 & 215 \\
Hospitality \& dining & 4 & 0 & 1 & 311 \\
Leisure \& entertainment & 4 & 1 & 1 & 199 \\
Commercial services & 2 & 1 & 1 & 57 \\
Institutional & 1 & 1 & 1 & 76 \\
Other environments & 1 & 1 & 1 & 77 \\
\midrule
\textbf{Total} & \textbf{24} & \textbf{5} & \textbf{8} & \textbf{1,155} \\
\bottomrule
\end{tabular*}
\vspace{0.5em}

\begin{tabular*}{\columnwidth}{@{\extracolsep{\fill}}lrrr@{}}
\toprule
Partition & Train & Val. & Test \\
\midrule
Recordings & 755 & 163 & 237 \\
Timestamps & 609,652 & 134,637 & 196,944 \\
\bottomrule
\end{tabular*}
\end{table}

\subsection{Ground-Truth Depth Construction}
\label{sec:gt_pipeline}

\begin{figure*}[t]
\centering
\includegraphics[width=\textwidth,trim={27bp 35bp 26bp 76bp},clip]{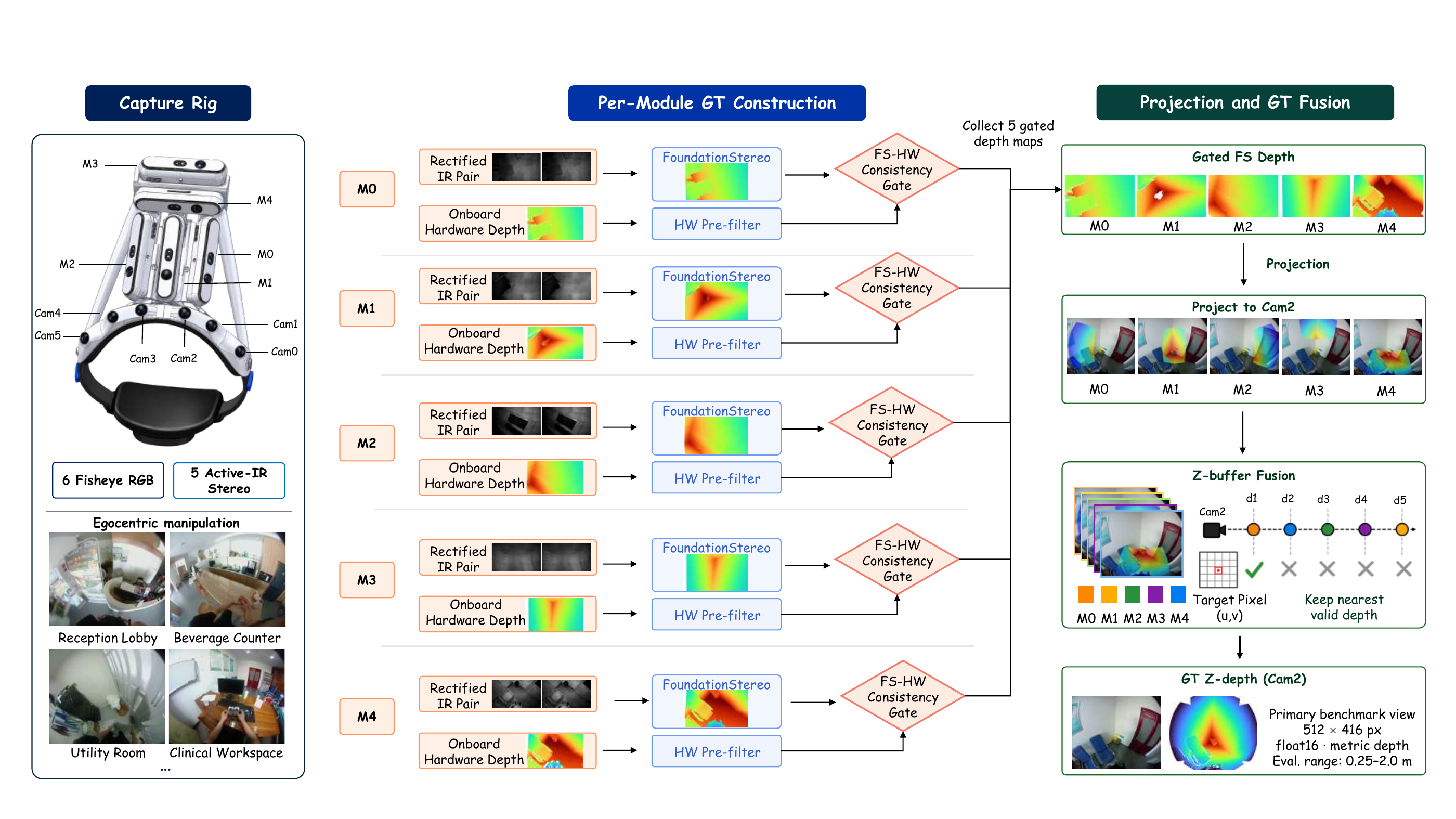}\par
\nointerlineskip
\setlength{\abovecaptionskip}{4pt}
\caption{Ground-truth depth construction. The rig records six fisheye RGB views
and five active-IR stereo pairs during manipulation. Per-module IR depth is
checked against filtered onboard depth, with a near-range exception where
onboard depth is unavailable, then projected and fused in camera~2.
The output is a $512\times416$ optical-axis Z-depth map aligned with camera~2's fisheye image.}
\label{fig:gt_pipeline}
\end{figure*}

The pipeline in Fig.~\ref{fig:gt_pipeline} reconstructs depth from five IR
viewpoints and fuses it in camera~2. IR pairs and onboard depth are matched
to camera~2 by timestamp; incomplete timestamps are skipped.
We apply FoundationStereo~\cite{wen2025foundationstereo} to rectified IR
pairs from the five Gemini~336L modules for smoother depth maps and
extended near-range coverage. For module $m$, the estimated disparity
$d_m$ yields metric depth
\begin{equation}
  z_m(p)=\frac{f_m B_m}{d_m(p)},
  \label{eq:ir_metric_depth}
\end{equation}
where $f_m$ is the rectified horizontal focal length in pixels and $B_m$
is the calibrated baseline in meters. We remove invalid correspondences
and small disparity components.

Onboard depth from the 336L's built-in algorithm provides a consistency
check. We filter isolated components and locally inconsistent background returns from both
onboard and learned depth. Where filtered onboard depth $h_m$ is valid,
learned depth must satisfy $|z_m-h_m|/h_m\leq0.05$.
Where onboard depth is invalid, only learned disparities $d_m>128$\,px in
the rectified IR images remain eligible, allowing reconstruction in the
onboard algorithm's near blind zone. All retained depth values come from
learned IR stereo.

We transform each retained source pixel $p$ into camera~2 and project it
into its fisheye image:
\begin{equation}
  \mathbf{P}_2=\mathbf{R}_{2m}\bigl(z_m(p)\mathbf{K}_m^{-1}\tilde p\bigr)
  +\mathbf{t}_{2m},\qquad q=\pi_2(\mathbf{P}_2),
  \label{eq:gt_projection}
\end{equation}
where $\tilde p$ is the homogeneous pixel, $\mathbf{K}_m$ contains rectified
intrinsics, and $(\mathbf{R}_{2m},\mathbf{t}_{2m})$ maps the left IR frame
to camera~2. The calibrated double-sphere model
$\pi_2$~\cite{usenko2018doublesphere} maps the transformed point to
image coordinates. A neighborhood visibility check uses camera-to-point range
to remove occluded background points. A z-buffer then retains the smallest
positive camera-2 Z at each pixel within a module, and fusion takes the
minimum across modules.

The fused output stores camera-2 optical-axis depth in meters, with zero
at unsupported pixels and a separate validity mask. The convex hull of
valid ground-truth pixels spans 83.1\% of the image area on average,
with depth measurements available at 66.9\% of all image pixels.

\subsection{Geometric Accuracy Validation}
\label{sec:validation}

\begin{figure}[t]
\centering
\includegraphics[width=\columnwidth,trim={1bp 12bp 0bp 78bp},clip]{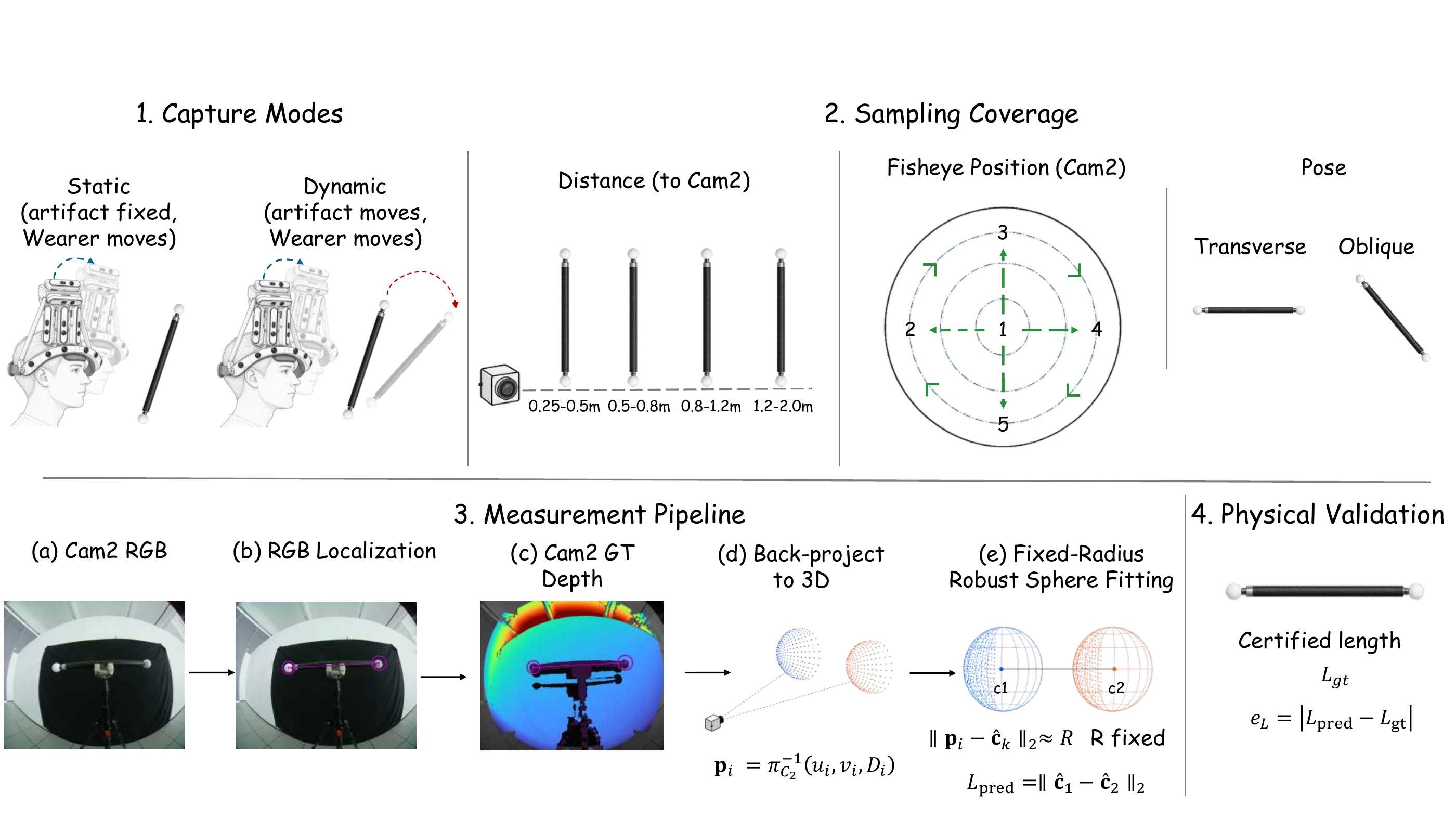}\par
\nointerlineskip
\setlength{\abovecaptionskip}{4pt}
\caption{Ball-bar validation across distances, viewing directions, and motion
conditions. Sphere pixels are back-projected and fitted with fixed radii;
the recovered center spacing is compared with the certified length.
Here $D_i$ is camera-2 optical-axis Z-depth, and $\pi_{C_2}^{-1}$ denotes
calibrated back-projection using this depth.}
\label{fig:ballbar_pipeline}
\end{figure}

We assess recovered dimensions using a ball-bar with certified center
spacing $L_{\rm gt}=499.9903\pm0.005$\,mm. Static and moving captures span
distances and fisheye positions, including orientations with the spheres
at different depths (Fig.~\ref{fig:ballbar_pipeline}). After localizing
sphere-pair candidates using appearance and depth checks, we back-project
their depth pixels and fit centers by Tukey-weighted least squares with
fixed certified radii (approximately 25\,mm).
The recovered centers $\hat{\mathbf{c}}_1,\hat{\mathbf{c}}_2$ give the spacing error
\begin{equation}
  e_L=\left|\|\hat{\mathbf{c}}_1-\hat{\mathbf{c}}_2\|-L_{\rm gt}\right|.
  \label{eq:ballbar_error}
\end{equation}
Sphere fits require at least 20 inliers and RMS $\leq4$\,mm;
between-sphere maximum/minimum ratios must be $\leq3$ for RMS and
$\leq10$ for inlier counts. These fit-quality checks yield 1,220
accepted measurements (38.8\% of localized frames) for accuracy analysis.

Accepted measurements have a median absolute center-spacing error of
1.27\,mm (P95: 6.66\,mm). Median error rises from 1.11\,mm at
0.25--0.8\,m to 9.79\,mm in the 1.2--2.0\,m bin
(Table~\ref{tab:ballbar_spatial}).

\begin{table}[t]
\caption{Absolute ball-bar spacing error, grouped separately by distance and
field angle. Distance is the range of the reconstructed midpoint; field angle
is its angle from the optical axis.}
\label{tab:ballbar_spatial}
\centering
\small
\begin{tabular*}{\columnwidth}{@{\extracolsep{\fill}}lrr@{}}
\toprule
Condition & Median (mm) & P95 (mm) \\
\midrule
\multicolumn{3}{l}{\textit{Distance (m)}} \\
$[0.25,\,0.8)$ & 1.11 & 4.57 \\
$[0.8,\,1.2)$ & 3.01 & 9.15 \\
$[1.2,\,2.0)$ & 9.79 & 22.78 \\
\midrule
\multicolumn{3}{l}{\textit{Field angle (degrees)}} \\
$[0,\,20)$ & 1.11 & 5.09 \\
$[20,\,40)$ & 1.50 & 10.20 \\
$[40,\,60)$ & 3.08 & 9.79 \\
\bottomrule
\end{tabular*}
\end{table}

\subsection{Data Curation and Splits}
\label{sec:data_splits}

We verify image and depth-map readability and dimensions, calibration, and
validity masks for every observation. Exact duplicate recordings are
removed by matching timestamps, RGB, depth, masks, and calibration.
We preserve temporal order and partial ground-truth coverage, without
filtering by image blur, scene difficulty, or model performance.

All recordings and synchronized views of each physical site, including
revisits, belong to one partition (Table~\ref{tab:dataset_statistics}).
Training and test sets each cover all seven environment groups.
The eight test sites comprise a fruit store, a gaming
arcade, a real-estate agency, a guesthouse, a horticultural garden, a
training center, a domestic dining room, and a living room.
Validation sites are used for checkpoint selection; test sites are held
out from fine-tuning.

The data package includes masks, calibrated rays, site assignments,
and manifests linking RGB and depth. Training and evaluation use fixed
temporal subsets of these partitions, as specified in
Section~\ref{sec:experimental_setup}. Sampling rules and evaluation code will accompany
the data.

\setlength{\textfloatsep}{9pt plus 2pt minus 2pt}
\setlength{\dbltextfloatsep}{9pt plus 2pt minus 2pt}
\section{Experiments}
\label{sec:experiments}

\subsection{Experimental Setup}
\label{sec:experimental_setup}

\begin{table*}[t]
  \caption{Public-checkpoint evaluation on eight test sites with equal weights.
  Full-image scores use the main test set; ROI entries give MAE$_Z$ (cm) /
  coverage (\%) on 291 hand and 257 object frames. Oracle rows align to target-frame
  ground truth; ``+ rig'' adds metric camera-to-rig poses.}
  \label{tab:main_results}
  \centering
  \footnotesize
  \setlength{\tabcolsep}{3pt}
  \begin{tabular*}{\textwidth}{@{\extracolsep{\fill}}lccccccc@{}}
    \toprule
    & & \multicolumn{4}{c}{Main test set}
      & \multicolumn{2}{c}{Annotated subset} \\
    \cmidrule(lr){3-6}\cmidrule(l){7-8}
    Method & Views & AbsRel$\downarrow$
      & \shortstack{MAE$_Z$\\(cm)$\downarrow$}
      & $\delta_1\uparrow$ & \shortstack{Coverage\\(\%)$\uparrow$}
      & \shortstack{Hand\\MAE$_Z\downarrow$/Cov.$\uparrow$}
      & \shortstack{Object\\MAE$_Z\downarrow$/Cov.$\uparrow$} \\
    \midrule
    \multicolumn{8}{l}{\textit{Metric monocular: camera 2; model-supported calibration; no alignment}} \\
    UniDepth v2~\cite{piccinelli2025unidepthv2} & 1 & 0.201 & 15.44 & 0.706 & 100.0 & 15.05/100.0 & 11.52/100.0 \\
    Depth Pro~\cite{bochkovskii2025depthpro} & 1 & 2.152 & 149.66 & 0.019 & 100.0 & 78.01/100.0 & 90.91/100.0 \\
    Metric3D v2~\cite{hu2024metric3dv2} & 1 & 0.452 & 37.37 & 0.296 & 100.0 & 17.95/100.0 & 20.66/100.0 \\
    DA3 Metric-Large~\cite{depth_anything_3} & 1 & 0.458 & 31.61 & 0.414 & 100.0 & 10.72/100.0 & 12.19/100.0 \\
    ZoeDepth~\cite{bhat2023zoedepth} & 1 & 2.457 & 159.24 & 0.006 & 100.0 & 115.11/100.0 & 116.82/100.0 \\
    UniK3D~\cite{piccinelli2025unik3d} & 1 & 0.392 & 30.63 & 0.387 & 100.0 & 33.77/100.0 & 31.33/100.0 \\
    DAC~\cite{guo2025dac} & 1 & 0.411 & 38.91 & 0.246 & 96.7 & 9.92/99.3 & 17.74/99.9 \\
    UniDAC~\cite{ganesan2026unidac} & 1 & 0.242 & 20.30 & 0.519 & 99.2 & 16.88/100.0 & 14.04/100.0 \\
    \midrule
    \multicolumn{8}{l}{\textit{Calibrated RGB stereo: cameras 2--3; rectified overlap; no alignment}} \\
    FoundationStereo~\cite{wen2025foundationstereo} & 2 & 0.068 & 7.35 & 0.951 & 35.4 & 2.15/72.8 & 2.52/81.0 \\
    GeoStereo~\cite{wei2026geostereo} & 2 & 0.078 & 8.24 & 0.954 & 35.4 & 2.28/72.8 & 2.75/81.0 \\
    \midrule
    \multicolumn{8}{l}{\textit{Metric reconstruction from RGB: no supplied rig poses; no alignment}} \\
    MASt3R~\cite{leroy2024mast3r} & 2 & 1.499 & 93.22 & 0.074 & 100.0 & 70.06/100.0 & 66.06/100.0 \\
    CUT3R~\cite{wang2025cut3r} & 6 & 1.100 & 64.58 & 0.184 & 100.0 & 64.17/100.0 & 62.93/100.0 \\
    \addlinespace[2pt]
    \multicolumn{8}{l}{\textit{Metric multi-view with calibration: synchronized RGB and per-camera rays; no alignment}} \\
    MapAnything~\cite{keetha2025mapanything} & 6 & 3.168 & 200.76 & 0.031 & 100.0 & 198.23/100.0 & 179.28/100.0 \\
    MapAnything + rig~\cite{keetha2025mapanything} & 6 & 0.708 & 49.07 & 0.204 & 100.0 & 36.51/100.0 & 33.81/100.0 \\
    \midrule
    \multicolumn{8}{l}{\textit{Oracle-aligned multi-view: synchronized RGB; scale-only alignment}} \\
    VGGT~\cite{wang2025vggt} & 6 & 0.428 & 30.59 & 0.365 & 100.0 & 21.90/100.0 & 16.83/100.0 \\
    VGGT-$\Omega$~\cite{wang2025vggtomega} & 6 & 0.284 & 19.60 & 0.553 & 100.0 & 10.43/100.0 & 12.66/100.0 \\
    \addlinespace[2pt]
    \multicolumn{8}{l}{\textit{Oracle-aligned monocular: camera 2; representation-specific alignment}} \\
    Depth Anything v2~\cite{yang2024depthanythingv2} & 1 & 0.247 & 20.55 & 0.533 & 100.0 & 6.57/100.0 & 11.51/100.0 \\
    Marigold~\cite{ke2024marigold} & 1 & 0.275 & 19.01 & 0.569 & 100.0 & 13.44/100.0 & 12.51/100.0 \\
    MoGe~\cite{wang2025moge} & 1 & 0.323 & 21.13 & 0.497 & 100.0 & 14.13/100.0 & 12.77/100.0 \\
    \bottomrule
  \end{tabular*}
\end{table*}

\noindent\textbf{Data and inputs.}
From the disjoint site partitions (Table~\ref{tab:dataset_statistics}), we sample
61,085 timestamps for training at 3\,Hz, 4,491 for validation at 1\,Hz,
and 6,574 for testing at 1\,Hz. At frequency $f$, times
$t_0+(k+0.5)/f$ are matched to the nearest eligible timestamp within 50\,ms,
with $t_0$ the recording start and ties favoring earlier frames.
Monocular models receive camera~2, stereo models cameras~2/3, and
multi-view models the listed synchronized views and supported calibration.
Predictions map to camera~2's $512\times416$ image coordinates,
preserving crop and rectification support.

We retain checkpoint preprocessing. Depth Pro estimates focal
length from RGB; ZoeDepth uses NYU weights. Metric3D and DA3 use camera-2
focal parameters at the evaluation resolution for metric scaling under a pinhole approximation.
UniDepth~v2 and UniK3D use an extended unified camera model (EUCM),
with maximum ray deviation $0.27^\circ$ across five rig calibrations.
DAC and UniDAC use double-sphere rays for equirectangular (ERP)
projection and back-projection. Stereo baselines, including FoundationStereo,
receive camera~2/3 RGB pairs rectified from the native fisheye views to
$800\times640$ at $90^\circ$ horizontal FoV.
MapAnything uses Apache weights and per-camera calibrated rays;
\textit{+ rig} adds metric camera-to-rig poses.
Other reconstruction conditions receive no poses. CUT3R resets state
per timestamp and processes camera~2 last to use all six views.

\noindent\textbf{Depth, support, and scale.}
We score optical-axis depth Z. Range converts as $z=r u_z$ for calibrated
unit ray $\mathbf{u}$; camera-frame point maps use their third coordinate.
Ground-truth domain $G$ contains valid pixels with $z^*\in[0.25,2.0]$\,m;
$P$ contains finite, positive predictions on actual output support.
AbsRel, MAE$_Z$ (cm), and $\delta_1$ at 1.25 use $G\cap P$;
coverage is $|G\cap P|/|G|$. Predictions are not clipped.
Scores cover the full image unless a region mask is specified.
Metric outputs remain unaligned. Separate oracle groups
fit least-squares positive Z scale for VGGT, VGGT-$\Omega$, and MoGe,
affine inverse depth for Depth Anything~v2, and affine Z for Marigold.
Each inference call uses one fit over full-target $G$ and
representation-valid output support, reused across regions.
Oracle scores use test ground truth and do not establish metric scale recovery.

\begin{figure*}[t]
  \centering
  \includegraphics[width=\textwidth]{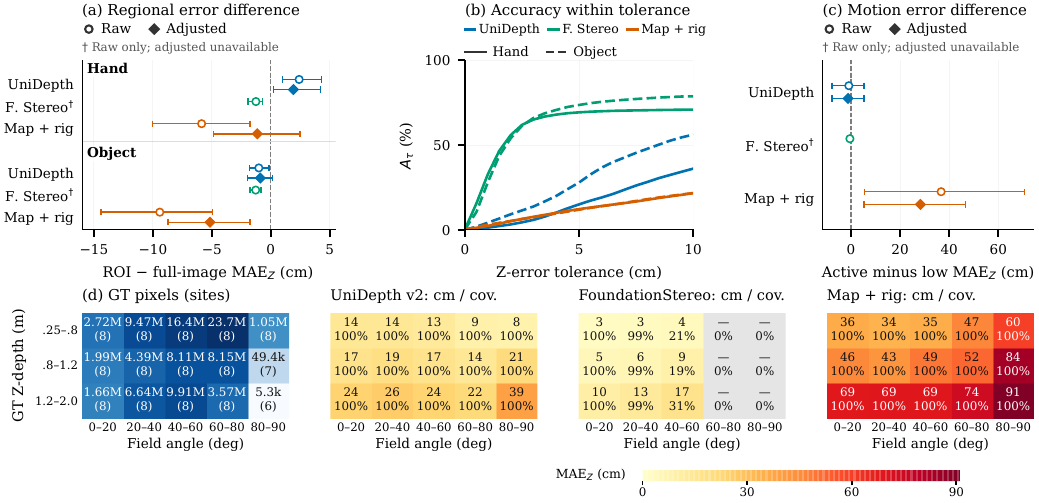}
  \par\nointerlineskip
  \setlength{\abovecaptionskip}{4pt}
  \caption{Interaction diagnostics for public checkpoints.
  (a) Regional-minus-full-image MAE at eight sites; negative values indicate lower regional error.
  (b) $A_\tau$: fraction of eligible GT pixels predicted within $\tau$; missing predictions fail.
  (c) Motion contrasts on nine shared cells at four sites.
  In (a,c), adjustment uses within-site GT Z--angle distributions;
  bars: 95\% site-bootstrap intervals. $\dagger$: missing cell predictions prevent adjustment.
  (d) Pooled GT counts (sites; log shading) and site-mean MAE$_Z$/coverage on 800 frames.
  MAE uses available predictions.}
  \label{fig:spatial_diagnostics}
\end{figure*}

\noindent\textbf{Aggregation.}
Pixels pool within sites; sites receive equal weight.
Paired 95\% intervals use 2,000 shared site bootstrap resamples;
specified model pairs also use common prediction support.
A site with ground truth but no predictions has zero coverage and undefined
error, leaving the complete error mean undefined.

\subsection{Zero-Shot Benchmark}
\label{sec:zero_shot}

Table~\ref{tab:main_results} compares 18 public models by input and scale
convention. UniDepth~v2 has the lowest metric monocular AbsRel (0.201)
and 15.44\,cm MAE with complete coverage.
Within its stereo overlap, FoundationStereo reaches 7.35\,cm MAE
with 35.4\% coverage.
On common support, its AbsRel is 0.068 versus 0.215 for UniDepth~v2,
a paired reduction of 0.146 [0.121, 0.178].

\subsection{Depth During Interaction}
\label{sec:diagnostics}

\noindent\textbf{Diagnostic subset.}
UniDepth~v2, FoundationStereo, and MapAnything + rig represent
monocular, stereo, and multi-view inputs.
Recording-balanced sampling selects 800 test samples (100 per site)
before annotation, independently of errors. RGB context at
$\pm0.5$ and $\pm1$\,s uses a 100\,ms tolerance. Qwen3-VL proposes
descriptions, Rex-Omni grounds objects, and SAM3 supplies
masks~\cite{qwen2025qwen3vl,jiang2025rexomni,carion2025sam3}.
We manually review RGB context and mask overlays, inspecting
ambiguous cases at native resolution.
Accepted hand/object masks contain valid ground truth in 291/257 frames
across eight sites each. Unaccepted regions remain unknown; the main test
set is unchanged. Ground truth covers 89.8\%/87.9\%
of accepted hand/object masks; 80.2\%/79.7\% fall within the Z interval (site means).

\noindent\textbf{Local accuracy and coverage.}
For region of interest (ROI) $R$, accuracy within Z-error tolerance $\tau$ is
\begin{equation}
 A_\tau(R)=
 \frac{\sum_{x\in G\cap R\cap P}
 \mathbf{1}[|\hat z(x)-z^*(x)|\leq\tau]}{|G\cap R|}.
 \label{eq:error_tolerance}
\end{equation}
Missing predictions count as failures. At $\tau=5$\,cm, object $A_\tau$ is
28.0\% for UniDepth~v2 (complete coverage) and 73.7\% for
FoundationStereo (81.0\% coverage).
Figure~\ref{fig:spatial_diagnostics}b reports depth accuracy over 0--10\,cm tolerances.

\noindent\textbf{Depth and viewing angle.}
To compare regions at similar geometry, we cross Z intervals
$[0.25,0.8)$, $[0.8,1.2)$, and $[1.2,2.0]$\,m with field angles
$\theta=\arccos u_z$ in $[0,20)$, $[20,40)$, $[40,60)$,
$[60,80)$, and $[80,90)$ degrees. Comparisons pair each ROI with the full image (including
the ROI) on frames with ROI ground truth. Site--Z--angle cells require at least
200 valid ROI depth pixels, retaining 99.6/99.8\% of hand/object ground-truth pixels.
Raw and adjusted differences share cells; adjustment weights
full-image errors by each ROI's within-site ground-truth distribution
(Fig.~\ref{fig:spatial_diagnostics}a).

UniDepth~v2's adjusted regional-minus-full-image MAE is
1.92 [0.21, 4.22]\,cm for hands and $-0.89$ [$-1.99$, 0.14]\,cm
for objects. MapAnything + rig's object MAE is 5.15 [1.75, 8.73]\,cm
below its matched full-image error, yet remains 33.81\,cm.

Figure~\ref{fig:spatial_diagnostics}d shows stereo's coverage gap beyond
$60^\circ$. UniDepth~v2's error does not increase uniformly with angle
at fixed Z; extreme-angle support is limited, especially at larger Z.

\begin{figure*}[t]
  \centering
  \includegraphics[width=\textwidth]{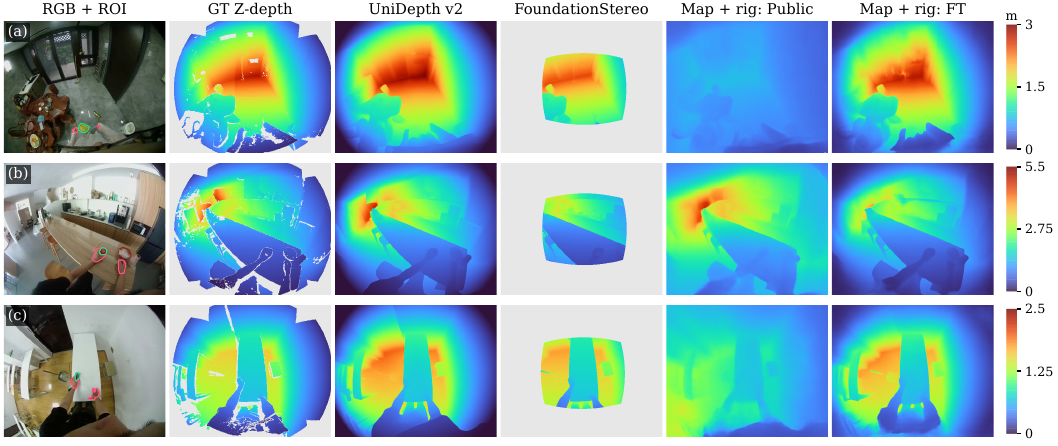}
  \par\nointerlineskip
  \setlength{\abovecaptionskip}{4pt}
  \caption{Qualitative depth at three test sites. Pink and green contours mark
  hands and objects. Each row shares a metric Z-depth scale; gray denotes
  invalid depth. Methods retain their own support. Public and FT share six
  RGB views, calibrated rays, and rig poses.}
  \label{fig:interaction_qualitative}
\end{figure*}

\noindent\textbf{Motion conditions.}
RGB context labels hand activity, object motion relative to the environment,
and background-image motion. Active clips show hand-driven object motion;
low-motion clips have little change in all three, allowing holding jitter.
Review of the fixed samples yields
59 active and 36 low-motion object frames across seven sites each.
Requiring 200 ground-truth pixels per condition retains nine of 50 occupied
site--Z--angle cells across four sites, all at $Z<0.8$\,m.
Raw and adjusted contrasts share these cells; adjustment uses within-site
pooled ground-truth distributions.

Adjusted active-minus-low differences are $-1.21$ [$-7.67$, 5.20]\,cm
for UniDepth~v2 and 28.29 [5.37, 46.57]\,cm for MapAnything + rig
(Fig.~\ref{fig:spatial_diagnostics}c). MapAnything without poses gives
$-30.17$ [$-73.24$, 6.21]\,cm. These contrasts are observational:
groups differ in objects and potentially in occlusion and appearance.

\subsection{Effects of Views and Rig Geometry}
\label{sec:additional_views}

With MapAnything's Apache checkpoint fixed, we separate view count from rig geometry:
camera~2, cameras~2/3, or all six views, each with calibrated rays.
A fourth adds six metric camera-to-rig poses
(Table~\ref{tab:view_ablation}). All omit depth hints
and optional confidence filtering.

\begin{table}[t]
  \caption{One Apache MapAnything checkpoint with 1/2/6 views and
  calibrated rays; ``+ rig'' also supplies metric poses. MAE$_Z$ is in cm. Full-image coverage is at least
  99.99\%; object coverage is 100\%. Cost is median forward time (ms)
  / peak allocated memory (GB), batch one, BF16, on RTX 4090.}
  \label{tab:view_ablation}
  \centering\footnotesize
  \setlength{\tabcolsep}{2pt}
  \begin{tabular*}{\columnwidth}{@{\extracolsep{\fill}}cccccc@{}}
    \toprule
    & \multicolumn{2}{c}{Full image} & \multicolumn{2}{c}{Objects} & \\
    \cmidrule(lr){2-3}\cmidrule(lr){4-5}
    Views & AbsRel$\downarrow$
      & MAE$_Z\downarrow$
      & MAE$_Z\downarrow$
      & $A_{5\mathrm{cm}}$ (\%)$\uparrow$
      & Cost \\
    \midrule
    1 & 2.800 & 176.70 & 173.19 & 0.01 & 58/7.7 \\
    2 & 3.392 & 211.34 & 188.00 & 0.03 & 88/8.2 \\
    6 & 3.168 & 200.76 & 179.28 & 0.01 & 233/10.1 \\
    6 + rig & 0.708 & 49.07 & 33.81 & 11.91 & 235/10.1 \\
    \bottomrule
  \end{tabular*}
\end{table}

Additional views give no clear accuracy gain: on common support,
six-minus-one AbsRel is 0.368 [$-0.191$, 1.036] across eight sites,
while median forward time rises from 58 to 233\,ms.

Rig poses primarily improve metric scale, reducing six-view AbsRel from 3.168 to 0.708.
Independent per-frame oracle scale fits on shared support give
0.334 without poses and 0.327 with poses,
a difference of $-0.008$ [$-0.014$, $-0.002$].
VGGT-$\Omega$ reaches 0.284 under the same oracle scale rule and shared
support.

\subsection{Adaptation to Held-Out Sites}
\label{sec:adaptation}

We assess adaptation on EgoDepth-Bench against global scale calibration. Public, Scale-only,
and fine-tuned (FT) models share
Apache initialization, six RGB views, calibrated rays, and metric rig poses.
Scale-only fits a positive multiplier to the frozen model on
1,024 unique training samples using frame-mean log-Z Huber loss
($\delta=0.1$). FT uses the same loss on camera~2 to update cross-view/ray
modules and dense/scale heads, with image and pose encoders frozen.

Training uses 6,000 AdamW steps, batch 36, learning rate $10^{-5}$,
weight decay 0.01, and 200 warmup steps followed by cosine decay.
We sample with replacement, uniformly over sites, recordings within
sites, and timestamps within recordings. Six RTX~4090 GPUs use BF16
and activation checkpointing at $518\times406$.
Validation at $512\times416$ every 500 steps selects step 6,000 by
lowest site-mean AbsRel; training and validation take 12.1 hours.
One seed (20260912) is used; site intervals do not measure seed variability.

\begin{table}[t]
  \caption{MapAnything adaptation (Apache weights) with six RGB views,
  rays and metric rig poses. Aligned rows fit scale per frame and arm on
  full-image common support, reusing the fit for objects.}
  \label{tab:adaptation}
  \centering\footnotesize
  \setlength{\tabcolsep}{2pt}
  \begin{tabular*}{\columnwidth}{@{\extracolsep{\fill}}lccc@{}}
    \toprule
    Evaluation & Public & Scale-only & Fine-tuned \\
    \midrule
    \multicolumn{4}{l}{\textit{Full image}} \\
    AbsRel$\downarrow$ & 0.708 & 0.601 & 0.091 \\
    MAE$_Z$ (cm)$\downarrow$ & 49.07 & 43.05 & 7.03 \\
    Coverage (\%)$\uparrow$ & 100.0 & 100.0 & 100.0 \\
    \midrule
    Hand MAE$_Z$ (cm)$\downarrow$ & 36.51 & 30.55 & 3.82 \\
    Object MAE$_Z$ (cm)$\downarrow$ & 33.81 & 29.05 & 5.21 \\
    Object $A_{5\mathrm{cm}}$ (\%)$\uparrow$ & 11.91 & 12.02 & 59.96 \\
    Object coverage (\%)$\uparrow$ & 100.0 & 100.0 & 100.0 \\
    Active-object $A_{5\mathrm{cm}}$ (\%)$\uparrow$ & 14.12 & 13.65 & 61.79 \\
    Active-object coverage (\%)$\uparrow$ & 100.0 & 100.0 & 100.0 \\
    \midrule
    Aligned AbsRel$\downarrow$ & 0.327 & 0.327 & 0.072 \\
    Aligned object MAE$_Z$ (cm)$\downarrow$ & 12.44 & 12.44 & 4.78 \\
    \bottomrule
  \end{tabular*}
\end{table}

Fine-tuning reduces AbsRel at all eight test sites (Table~\ref{tab:adaptation}).
Training-fitted scale $s_C=0.88071$ lowers test AbsRel from
0.708 to 0.601; FT reaches 0.091. On common support,
FT-minus-Public AbsRel is $-0.617$ [$-0.802$, $-0.451$].
Object MAE falls from 33.81 to 29.05\,cm with scale calibration
and 5.21\,cm with FT, all at full coverage.

Relative depth also improves. Independent positive per-frame
scale fits on full-image common test support, reused for objects, give both
controls aligned AbsRel 0.327 and object MAE 12.44\,cm; FT reduces these to
0.072 and 4.78\,cm. Figure~\ref{fig:interaction_qualitative} shows three test examples.
Local errors persist: FT places 60.0\% of eligible object pixels within 5\,cm.

\section{Conclusion}
\label{sec:conclusion}

EgoDepth-Bench benchmarks metric depth during real egocentric manipulation
using six synchronized RGB views and native fisheye ground truth, with
ball-bar validation of recovered dimensions. Shared observations and separate
training and test sites support model comparisons and adaptation. Hand and
object diagnostics show that low local error can coexist with incomplete
prediction coverage.

The current protocol scores individual frames. Extending it to temporal
depth consistency would use the synchronized sequences to assess predicted
geometry through motion and visibility changes during interaction.

\IEEEtriggeratref{20}

\end{document}